\documentclass[preprint,12pt,authoryear]{elsarticle}
\usepackage[T1]{fontenc}
\usepackage[utf8]{inputenc}
\usepackage{lmodern}
\usepackage[margin=1in]{geometry}  % elsarticle's preprint text block is narrow
\usepackage{amsmath,amssymb,bm}
\usepackage{graphicx}
\usepackage{float}  % [H]: place a figure exactly where it appears in the text
\usepackage{xcolor}
\usepackage{hyperref}

\begin{document}

\begin{frontmatter}

\title{From Visual Search to Movement Control: A Priority Field for Artificial Agents}

\author[umtri]{Han Zhang}
\ead{hanzh@umich.edu}
\author[cee]{Zhong Cao}
\ead{zhcao@umich.edu}
% \cortext/\corref omitted: newer LaTeXML (arXiv HTML) renders the note as an affiliation.
% For journal submission, restore \corref{cor1} on the author and \cortext[cor1]{Corresponding author}.

\address[umtri]{University of Michigan Transportation Research Institute}
\address[cee]{Civil and Environmental Engineering, University of Michigan}

\begin{abstract}
Human spatial attention is widely conceptualized as being guided by a priority map that integrates perceptual salience, current goals, and past experiences. Here, we extend priority-based computation to movement control in artificial agents. We first introduce a lightweight model of visual search based on an integrated priority map. Trained on human fixations, it reproduced key behavioral patterns, including oculomotor suppression and history-driven selection. Extending the search model, we equipped an artificial agent with a priority field and evaluated its performance in a reach--avoid task that required reaching a goal destination while avoiding moving obstacles. Compared with alternative architectures, priority-field agents trained more efficiently and performed better in unseen, complex scenarios, even from simple demonstrations. Adding a simple memory mechanism also produced human-like, history-driven effects in anticipating the likely location of the upcoming goal. These findings suggest that priority-based computation may provide a promising foundation for movement control in artificial agents.
\end{abstract}

\end{frontmatter}

\section{Introduction}\label{sec:intro}

Deciding where to look is one of the most fundamental decisions humans make. The inherent biological constraints of the human visual system require us to prioritize certain visual information over others to survive and function efficiently \citep{desimone_neural_1995,carrasco_visual_2011}. This process is widely thought to arise from a dynamic interplay among bottom-up salience (e.g., which objects stand out to the observer), top-down goals (e.g., what the observer is looking for), and past experiences (e.g., where the target was previously located; \citealp{awh_top-down_2012,theeuwes_goal-driven_2019}). These sources of information converge in a ``priority map,'' which can be understood as a real-time representation of the significance of objects in the visual field \citep{fecteau_salience_2006,bisley_attention_2010,zelinsky_what_2015}. Higher values on this map signal a greater need for behavioral prioritization, including through eye movements \citep{fecteau_salience_2006,itti_computational_2001}.

Deciding where to move is not limited to eye movements. As we navigate a crowded dining hall, for example, we must watch our step to avoid bumping into others. An artificial agent, such as a self-driving car, must likewise ensure that its movements do not result in a collision. Fundamentally, controlling movements requires an internal representation of surrounding objects that signals the need for behavioral prioritization. Like a priority map, this internal representation should incorporate bottom-up salience, such as the locations of obstacles and how quickly they are approaching, as well as top-down goals, such as the intended destination \citep{fajen_behavioral_2003}. It may also incorporate past experiences, including the locations of previous destinations, to optimize behavior in response to environmental regularities. In other words, the cognitive principles that govern human eye movements may generalize to movement across space.

\begin{figure}[H]
  \centering
  \includegraphics[width=\linewidth, alt={Two-panel conceptual diagram. Panel A: in visual search, salience, goal, and history maps over item locations are summed into a priority map that guides the first fixation. Panel B: for agent movement, salience, goal, and history fields over movement directions are summed into a priority field that guides movement.}]{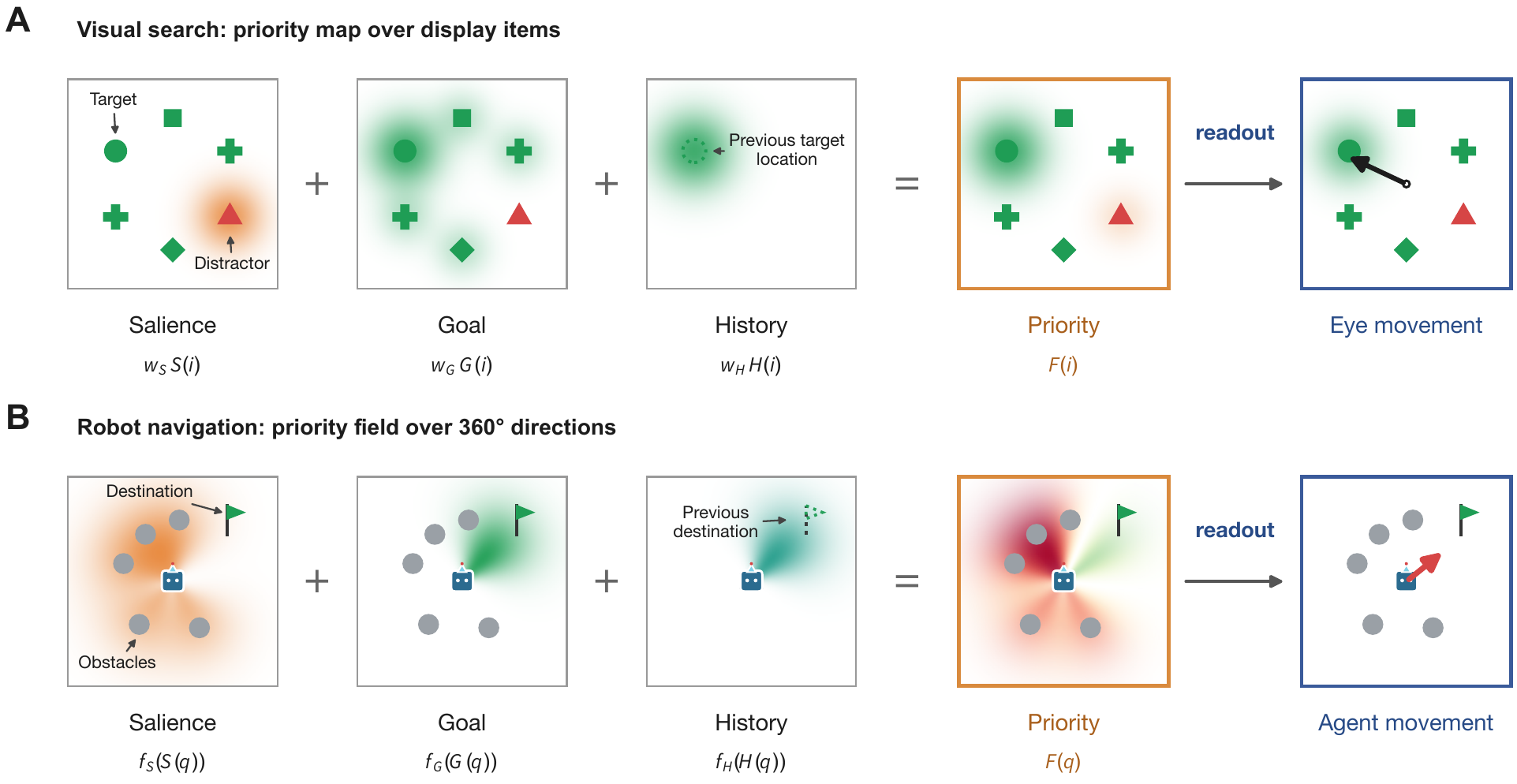}
  \caption{\textbf{Conceptual diagrams of priority-based models for visual search (A) and agent movement (B).} In both cases, behavior is guided by a priority representation that integrates perceptual salience, the current goal, and selection history. This representation is a priority map defined over item locations in visual search, and a priority field defined over movement directions in agent movement.}
  \label{fig:concept}
\end{figure}

To demonstrate this generalizability, we start by modeling human eye movements during visual search. Consider a classic feature-search task (see \autoref{fig:concept}A), in which participants search for a predefined target (e.g., a green circle) among heterogeneous shapes, with its location randomized across trials \citep{gaspelin_suppression_2017}. On some trials, a nontarget appears in a unique color (e.g., red), thus becoming a salient singleton distractor. Despite its salience, observers can suppress this distractor: first fixations are less likely to land on the singleton than on an average non-singleton distractor \citep{gaspelin_suppression_2017,gaspelin_role_2018}. Observers also show target-location priming, with first fixations biased toward the previous target location \citep{maljkovic_priming_1996}. This history-driven effect indicates an automatic tendency to optimize visual selection based on environmental regularities \citep{awh_top-down_2012,theeuwes_goal-driven_2019}.

A conceptual diagram of the visual-search model is shown in \autoref{fig:concept}A. The model adds three sources of information into a single priority map. For each candidate item $i$,
\begin{equation}\label{eq:search-general}
F(i) = f_{S}(S(i)) + f_{G}(G(i)) + f_{H}(H(i))
\end{equation}
where the salience term $S(i)$ indicates how strongly the item stands out perceptually, the goal term $G(i)$ indicates how well it matches the target-defining features, and the history term $H(i)$ indicates the influence of previous target locations. Each function $f$ maps a source into its contribution to the priority map. In the search model, for simplicity, we assume the mapping functions are scalar weights:
\begin{equation}\label{eq:search-linear}
F(i) = w_{S}S(i) + w_{G}G(i) + w_{H}H(i)
\end{equation}
The priority map is thus a weighted sum of three individual maps.

The visual-search model provides a useful basis for understanding spatial movements. Consider a robot navigating toward a destination while avoiding obstacles. Whereas human observers must avoid attending to the distractor to find the search target, the robot must avoid colliding with the obstacles to reach the goal. Whereas human observers anticipate upcoming target locations based on previous locations, the robot may anticipate upcoming destinations based on previous destinations. Thus, although visual search and spatial navigation differ in modality, they can be understood by the same set of mechanistic principles.

We therefore propose a model in which an agent's movement in space is guided by a priority field that encodes the influence of salience, goal, and history for each movement direction. A conceptual diagram is presented in \autoref{fig:concept}B. For each candidate direction $q$,
\begin{equation}\label{eq:field-general}
F(q) = f_{S}(S(q)) + f_{G}(G(q)) + f_{H}(H(q))
\end{equation}
where $S(q)$ reflects collision threat in that direction, $G(q)$ reflects alignment with the destination, and $H(q)$ reflects the influence of recent destinations. To maximize performance in a dynamic setting, the agent model uses small neural networks for its component functions. Summing its salience, goal, and history yields a priority field defined over possible movement directions. A small planner network converts this field into a movement, combining the pull of high-priority directions (e.g., toward the goal) with the push of low-priority directions (e.g., away from obstacles).

In the present study, we first trained the visual-search model on human eye movements and showed that it reproduced key behavioral patterns, including oculomotor suppression and history-driven effects. We then extended the model to movement control, equipping an artificial agent with a priority field as a foundational capability. To evaluate its performance, we used a reach--avoid task in which the agent needs to reach a goal location while avoiding moving obstacles. We also compared it with two architectures that lack an explicit priority-based computation: a multilayer perceptron (MLP) and a transformer. We found that priority-field agents trained more efficiently and generalized better to unseen, complex scenarios, even from simple demonstrations. They could also exhibit human-like history-driven effects in anticipation of the likely location of an upcoming goal.

\section{Results}\label{sec:results}

\subsection{A priority map for visual search}\label{sec:res-search}

We fit the visual search model to a public dataset containing 114,232 first fixations from 333 participants across 11 feature-search experiments \citep{drennan_what_2026}. On each trial, observers searched for a fixed target (e.g., a green circle) among heterogeneous distractor shapes. A color singleton distractor appeared at a random location on some trials (\autoref{fig:concept}A). The landing position of the first fixation on each trial was used, as it provides a direct readout of the observer's attentional priority over the search display \citep{gaspelin_role_2018}. The model was trained on 80\% of participants, leaving 20\% of participants (66 of 333) as the held-out test set.

Estimated model parameters align with theoretical accounts of search behavior \citep{wolfe_guided_2021,awh_top-down_2012}: $w_{S} = -0.05$, $w_{G} = 1.34$, $w_{H} = 2.09$, and $\eta_{H} = 0.59$. The near-zero salience weight indicates little influence of salience on fixation behavior, whereas the positive goal weight indicates prioritization of target-matching features. The history parameters together indicate a recency-weighted influence of previous target locations. On held-out participants, the model achieved a mean negative log-likelihood of 1.40, outperforming a human-consistency baseline of 1.54 (see Methods). This difference was statistically significant, $t(65) = 6.72$, $p < .001$, $d = 0.62$.

\begin{figure}[H]
  \centering
  \includegraphics[width=\linewidth, alt={Three bar-chart panels. A: singleton versus non-singleton distractor landing rates for held-out participants and model predictions, showing suppression of the singleton. B: model-predicted landing rates for high- versus low-salience singleton distractors. C: target-directed first fixations for repeated versus changed target locations in participants and the model.}]{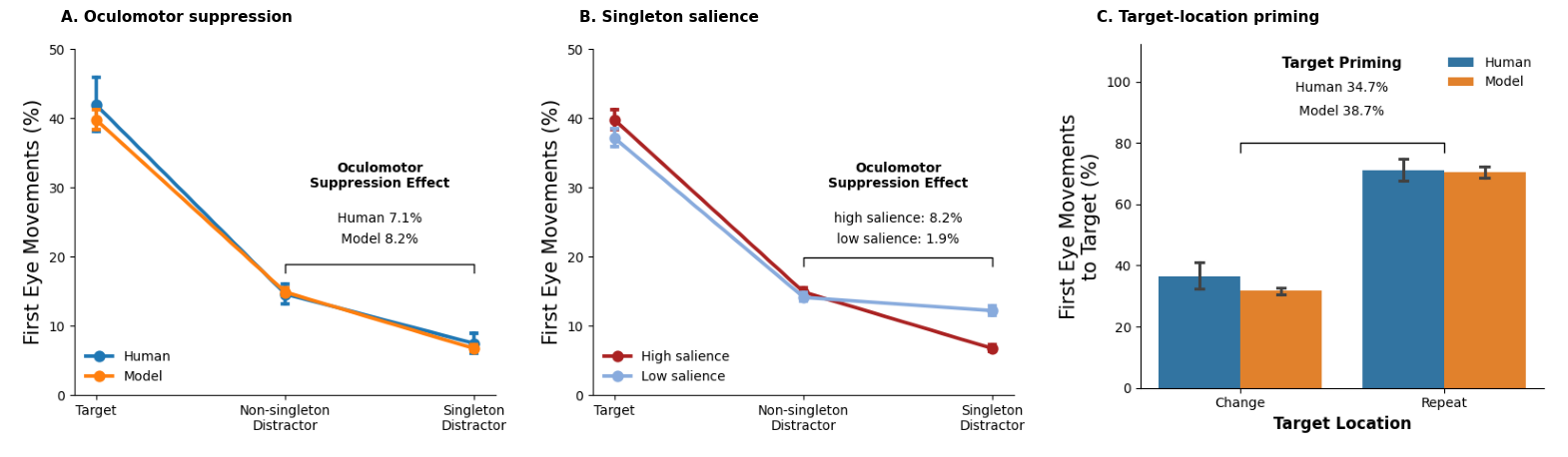}
  \caption{\textbf{The fitted model reproduced key patterns of visual search.} (A) Oculomotor suppression in held-out participants and model predictions. (B) Model predictions for high- and low-salience distractors. (C) Target-location priming in held-out participants and model predictions.}
  \label{fig:search}
\end{figure}

The fitted model also reproduced key signatures of human search observed in the literature \citep{drennan_what_2026}. First, it reproduced oculomotor suppression (\autoref{fig:search}A). Held-out participants fixated on the singleton distractor on 7.5\% of trials versus 14.6\% for an average non-singleton distractor, corresponding to a 7.1\% oculomotor suppression effect, $t(65) = 10.79$, $p < .001$, $d = 1.21$. On the same held-out participants, the model predicted 6.7\% versus 14.9\%, producing an 8.2\% oculomotor suppression effect, $t(65) = 23.35$, $p < .001$, $d = 3.68$.

Second, the model reproduced previous findings showing that suppression increases with distractor salience in this task (\autoref{fig:search}B; \citealp{stilwell_role_2023}). We regenerated predictions for the held-out distractor-present trials assuming a low-salience singleton distractor. Compared to the original predictions, the oculomotor suppression effect decreased from 8.2\% to 1.9\% for low-salience distractors. This is driven by the model assigning a higher landing probability to low-salience distractors, $t(65) = 15.07$, $p < .001$, $d = 2.30$.

Third, the model reproduced target-location priming (\autoref{fig:search}C; \citealp{maljkovic_priming_1996}). For each held-out participant, we compared target-directed first fixations when the target location was repeated versus when it changed. Human participants produced a 34.7\% priming effect ($t(65) = 18.55$, $p < .001$, $d = 2.11$). The model produced a similar 38.7\% priming effect, $t(65) = 45.86$, $p < .001$, $d = 6.31$.

\subsection{A priority field for agent movements}\label{sec:res-field}

\begin{figure}[H]
  \centering
  \includegraphics[width=0.85\linewidth, alt={Four arena snapshots. A: training scenario with 10 obstacles. B: 50 obstacles. C: 10 obstacles at four times the training speed. D: statistical-learning scenario with the left goal region marked. In each, colored rays around the agent show its priority field, green for high-priority and red for low-priority directions.}]{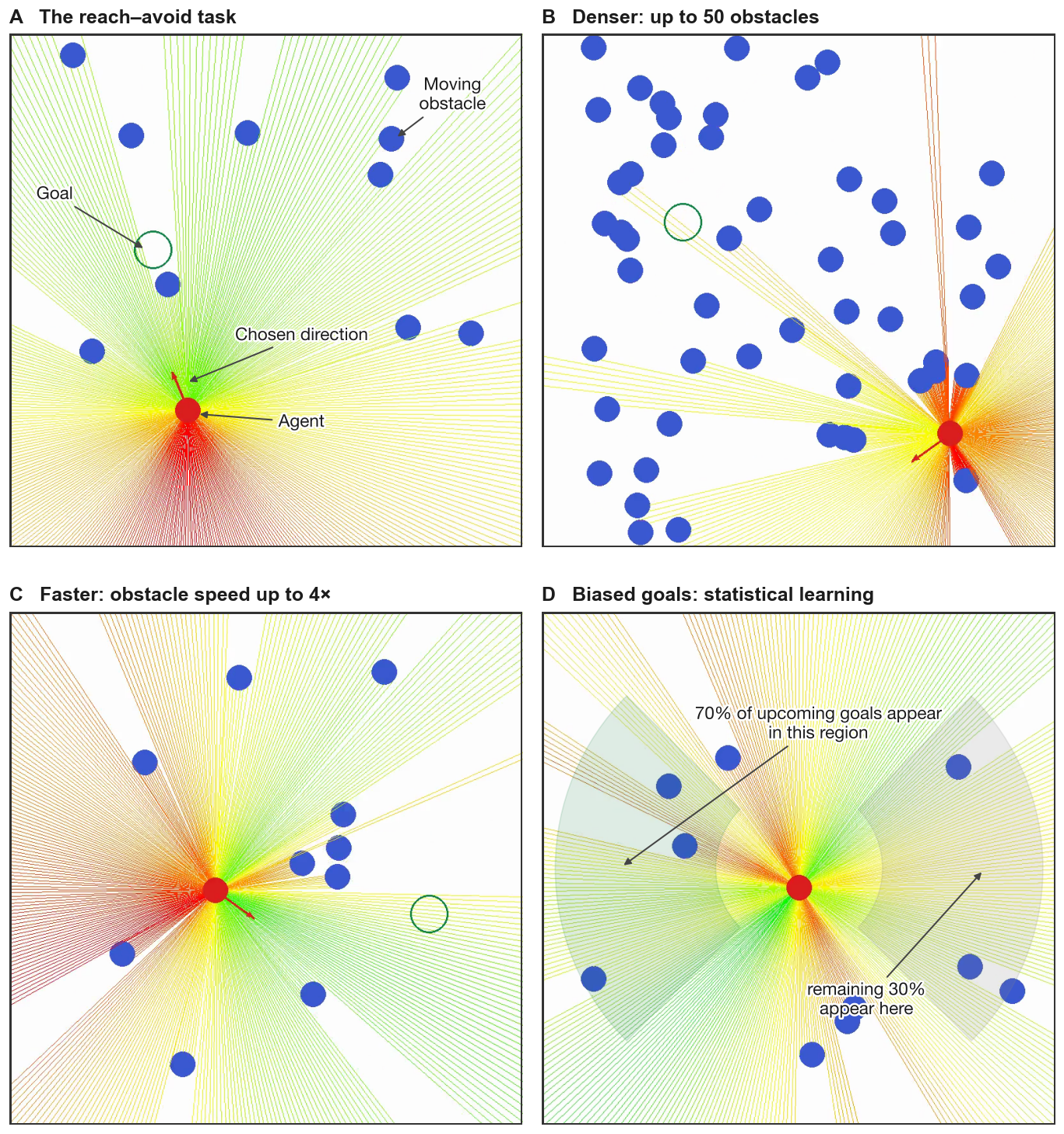}
  \caption{\textbf{Testing scenarios.} The agent's task is to reach the goal (green ring) while avoiding moving obstacles (blue circles). Panel A shows the demonstrated task used at training. Panel B shows one scalability test with 50 obstacles. Panel C shows another scalability test with obstacles moving at 4$\times$ speed. Panel D shows the statistical learning test, where the goal is more likely to appear inside the left region. In all panels, the agent's real-time priority field is shown as colored rays in each direction. The agent's movement reflects the combined pull of high-priority (green) directions and push of low-priority (red) directions.}
  \label{fig:scenarios}
\end{figure}

Having established that the priority-map model captures human eye movements during visual search, we asked whether the principles of priority-based guidance generalize to movement control. We equipped an artificial agent with a priority field and evaluated its performance in a reach--avoid task (see \autoref{fig:scenarios}A). In this task, the agent operated continuously in an $800\times800$-pixel arena containing moving obstacles and a single goal. The agent's task was to reach the goal while avoiding the moving obstacles. At each step, the agent sensed the distance to the nearest obstacle or the arena wall in every direction. The goal destination changed to a new location once the agent reached it or after 6 seconds. The training scenario contained 10 obstacles, each moving at 50--150 pixels per second along the horizontal and vertical axes.

The agent was trained via behavioral cloning from a scripted expert who demonstrated the task over 20 episodes (see Methods). Because the expert was memoryless and its demonstrations therefore contained no history-driven behavior, we fixed $w_{H} = 0$ during training and later set $w_{H}$ and $\eta_{H}$ separately (see Statistical Learning). For comparison, we trained two alternative architectures that received identical input but lacked explicit priority-based computation: a multilayer perceptron (MLP) and a transformer.

\begin{figure}[H]
  \centering
  \includegraphics[width=\linewidth, alt={Multi-panel comparison of the priority-field, MLP, and transformer agents. A: training loss curves on a log scale over 500 epochs. B: bar charts of goals and collisions per minute on the demonstrated task, with the expert's goal rate as a dashed line. C: goals and collisions per minute as obstacle count rises from 10 to 50. D: goals and collisions per minute as obstacle speed rises to four times the training level.}]{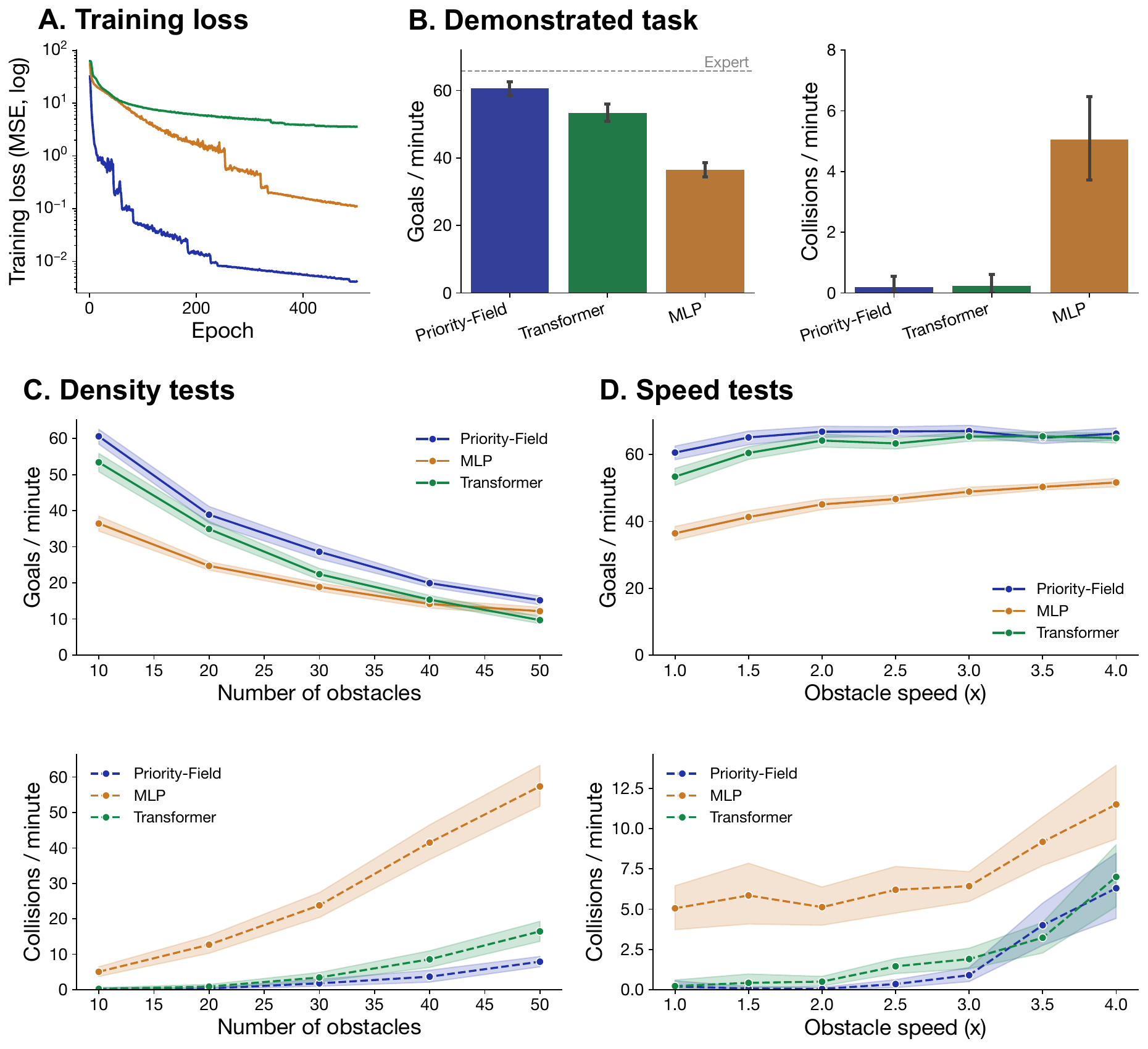}
  \caption{\textbf{Comparison of the priority-field agent against two alternative models without explicit priority-based computation (MLP, transformer).} (A) Training loss (mean squared error, log scale) over 500 epochs. (B) Performance on the demonstrated task (10 obstacles at the training speed): goals reached per minute (left; dashed line = scripted expert) and collisions per minute (right). (C) Density test: goals (top) and collisions (bottom) per minute as the number of obstacles increases from 10 to 50, with speed held at the training level. (D) Speed test: goals (top) and collisions (bottom) per minute as obstacle speed increases up to 4$\times$ the training level, with the number of obstacles held at 10. In B--D, bars/points show the mean across 20 random seeds, and error bars/shaded bands indicate 95\% bootstrap confidence intervals.}
  \label{fig:comparison}
\end{figure}

\paragraph{Training efficiency} The priority-field agent learned the expert policy more efficiently than alternative models (see \autoref{fig:comparison}A). With 500 training epochs, the priority-field agent reached a final mean squared error of 0.004. In comparison, the MLP agent and the transformer agent reached a final training loss of 0.111 and 3.593, respectively. The priority-field agent thus recovered the demonstrated behavior more quickly and accurately.

\paragraph{Demonstrated task} We first evaluated the agents' performance on the demonstrated task during training (10 obstacles at training speed). Performance is defined in terms of both speed (goals/minute) and safety (collisions/minute). The results are shown in \autoref{fig:comparison}B. The priority-field agent reached 60.60 goals/minute (95\% Confidence Interval: [58.50, 62.62]) while maintaining only 0.20 collisions/minute (95\% CI: [0.00, 0.55]). In comparison, the MLP agent reached 36.45 goals/minute (95\% CI: [34.30, 38.58]) while having 5.05 collisions/minute (95\% CI: [3.73, 6.47]). The transformer agent reached 53.40 goals/minute (95\% CI: [50.85, 55.95]) with 0.23 collisions/minute (95\% CI: [0.00, 0.60]). The priority-field agent reached statistically more goals than both the MLP ($t(19) = 21.63$, $p < .001$, $d = 4.93$) and the transformer ($t(19) = 6.17$, $p < .001$, $d = 1.33$). It also had statistically fewer collisions than the MLP ($t(19) = -7.44$, $p < .001$, $d = 2.06$), though it was indistinguishable from the transformer agent ($t(19) = -0.11$, $p = .92$, $d = 0.03$).

\paragraph{Scalability tests} Next, we evaluated performance in unseen and more challenging scenarios. First, we increased the number of obstacles in the scenario to up to 50, while holding their speed at the training level (see \autoref{fig:scenarios}B). The priority-field agent reached the most goals at every density and kept collisions lowest (see \autoref{fig:comparison}C). At the hardest level (50 obstacles), it reached 15.20 goals/minute (95\% CI: [13.93, 16.38]), which is statistically higher than the MLP (12.18, 95\% CI: [11.00, 13.30]; $t(19) = 4.22$, $p < .001$, $d = 1.10$) and the transformer (9.70, 95\% CI: [8.72, 10.68]; $t(19) = 9.98$, $p < .001$, $d = 2.13$). For collision rates, the priority-field agent resulted in 7.90 collisions/minute (95\% CI: [6.47, 9.40]) at the hardest level, compared to 57.38 (95\% CI: [51.83, 63.53]) for the MLP ($t(19) = -16.12$, $p < .001$, $d = 4.92$) and 16.45 (95\% CI: [13.57, 19.38]) for the transformer ($t(19) = -4.32$, $p < .001$, $d = 1.60$).

In the second test, we increased obstacle speed up to 4$\times$ the training speed while keeping the number of obstacles at 10 (see \autoref{fig:scenarios}C). As shown in \autoref{fig:comparison}D, the priority-field agent maintained a high goal rate ($\approx$65--67 goals/minute across elevated speeds) while collisions rose modestly. At 4$\times$ speed, it reached more goals than the MLP (66.25 [64.38, 68.00] vs.\ 51.65 [50.33, 52.92]; $t(19) = 12.69$, $p < .001$, $d = 3.98$) while remaining statistically indistinguishable from the transformer (64.95 [63.45, 66.40]; $t(19) = 1.52$, $p = .15$, $d = 0.34$). For collision rates, the priority-field agent collided less often than the MLP (6.30 [4.40, 8.57] vs.\ 11.50 [9.35, 14.00]; $t(19) = -4.22$, $p < .001$, $d = 1.01$) at 4$\times$ speed, while remaining statistically indistinguishable from the transformer (7.00 [5.15, 9.05]; $t(19) = -0.84$, $p = .41$, $d = 0.15$).

Overall, the priority-field agent trained more efficiently and generalized better to unseen scenarios than alternative architectures.

\paragraph{Statistical learning} Finally, we tested whether adding memory (enabling the history field) lets the agent adjust its behavior in response to environmental regularities \citep{geng_spatial_2005}. We used a scenario in which the goal location is spatially predictable (see \autoref{fig:scenarios}D). Each trial began with the agent teleported to the center of the arena, with no visible goal. After a 50-step goal-free period, a goal appeared in one of two symmetric wedges to the left or right of the arena center. A trial ended when the goal was reached or after 6 seconds. The agent completed a biased block of 90 trials, immediately followed by an unbiased block of 90 trials. During the biased block, the goal fell in the left region 70\% of the time. During the unbiased block, the goal had an equal probability of appearing in either region. We equipped the trained agent with memory of previous goal locations ($w_{H} = 0.20$, $\eta_{H} = 0.15$) and measured its anticipatory movement during the goal-free period. The dependent measure is the horizontal offset from the center at the moment when the goal appears. A negative offset, therefore, indicates that the agent drifted toward the high-probability region in anticipation of the upcoming goal. We compare the memory agent against an otherwise identical agent with memory disabled ($w_{H} = 0$) across 20 runs.

\begin{figure}[H]
  \centering
  \includegraphics[width=\linewidth, alt={Line plot of the agent's horizontal offset at goal onset across 180 trials in 10-trial bins. The agent with memory drifts to about minus 40 pixels during the first 90 biased trials and returns to near zero during the next 90 unbiased trials; the agent without memory stays near zero throughout.}]{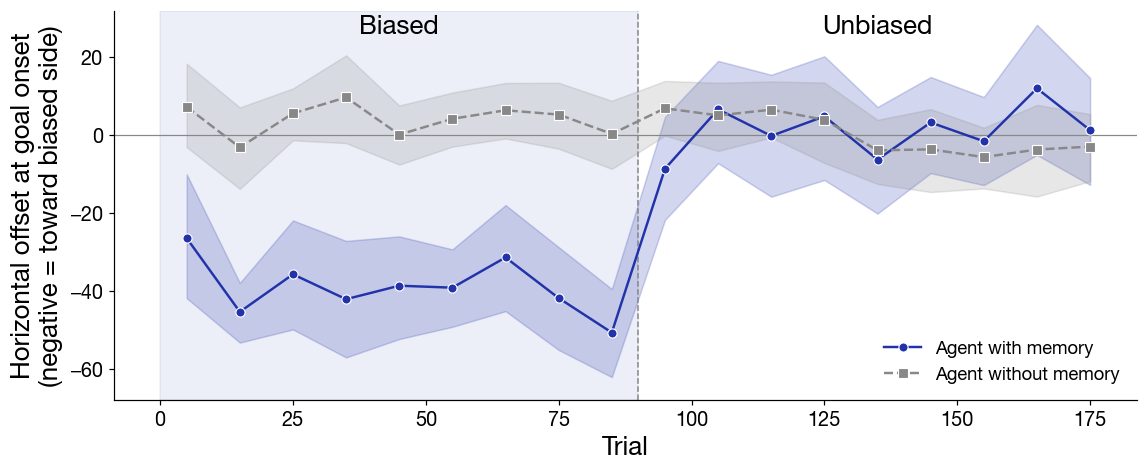}
  \caption{\textbf{The history field produces anticipatory, history-driven movement.} Anticipatory movement in the statistical-learning task, measured as the agent's horizontal offset from the arena center at goal onset (negative = toward the high-probability left region). Goals were spatially biased for the first 90 trials (70\% in the left region) and unbiased for the next 90 trials. ``Agent with memory'' is the priority-field agent with its history field enabled (strength $w_{H} = 0.20$, update rate $\eta_{H} = 0.15$). ``Agent without memory'' is the same agent with the history field disabled ($w_{H} = 0$). The memory agent drifted toward the high-probability region during the biased block and returned to the center once goals became unbiased. Each dot represents the mean offset over 10-trial bins. Shaded bands are bootstrap 95\% confidence intervals computed from 20 runs.}
  \label{fig:statlearn}
\end{figure}

The results are shown in \autoref{fig:statlearn}. During the biased block, the memory agent drifted toward the high-probability region with a mean offset of $-39.15$ pixels (95\% CI: [$-43.87$, $-34.74$]), compared with 3.96 pixels (95\% CI: [0.30, 7.92]) for the memoryless control ($t(19) = -16.48$, $p < .001$, $d = 4.37$). This history-driven effect extinguished in the unbiased block, during which the memory agent's offset became indistinguishable from the memoryless control (memory 1.28 pixels, 95\% CI [$-5.92$, 8.10] vs.\ control 0.32 pixels, 95\% CI [$-3.77$, 4.31]; $t(19) = 0.30$, $p = .76$, $d = 0.07$). These results show that the memory agent exhibited a human-like, history-driven effect, flexibly adjusting its behavior based on environmental regularities \citep{geng_spatial_2005}.

\section{Discussion}\label{sec:discussion}

In this work, we asked whether the priority-based computation used to explain human spatial attention can be used to study movement control. A lightweight priority-map model, fit to human first fixations, reproduced key signatures of visual search, including oculomotor suppression, salience dependence, and target-location priming. Extending the model to a priority field over movement directions, the same priority-based computation enabled an artificial agent to learn a reach--avoid task efficiently, handle unseen and challenging scenarios well, and produce human-like, history-driven effects in response to environmental regularities. These results suggest that deciding where to look and deciding where to move are seemingly different problems that share the same underlying structure.

The priority-field agent outperformed two alternative models that lack an explicit priority-based computation. By committing in advance to what must be represented (i.e., salience, goal, and history) while learning only how each source maps to behavior, the agent recovers behavior from little data and generalizes to unseen cases better than architectures without priority-based computation. This speaks, if only to a limited extent, to the imitation learning used to train autonomous vehicles, where structure-free policies are data-hungry and often difficult to interpret \citep{codevilla_exploring_2019,rudin_stop_2019}.

The history field in the artificial agent is implemented as a simple leaky accumulator, a mechanism widely used in models of human cognition \citep{usher_time_2001}. The fact that the same type of implementation yields history-driven effects in both visual search and agent movement connects movement control to the broader literature on experience-driven cognition, in which behavior is shaped automatically by past events, including statistical learning \citep{geng_spatial_2005,wang_statistical_2018}, contextual cueing \citep{chun_contextual_1998}, and reward \citep{anderson_value-driven_2011}. Furthermore, while we demonstrated the priority-field idea with an artificial agent, the same modeling approach can be used to study biological movement. Because the model is fit directly to behavior, it offers a way to quantify how salience, goal, and history shape movement in humans and other animals, ranging from visually guided reaching \citep{cisek_cortical_2007} to foraging \citep{wolfe_when_2013} and spatial navigation \citep{fajen_behavioral_2003}.

Several limitations qualify these conclusions. The search model was intentionally minimal, capturing only the mechanisms needed to explain behavior in the feature-search task. It does not separately estimate the upweighting of target features and the downweighting of distractor features and may not generalize to other search tasks. As a proof of concept, the reach--avoid task was limited to a two-dimensional simulation with range-only sensing and simple obstacles. Further development is needed to support richer perception and transfer to physical systems. Finally, the history field used a simple memory mechanism that tracked only goal locations; a more comprehensive account might also incorporate reward-based learning \citep{anderson_value-driven_2011,gottlieb_attention_2014}. Nevertheless, these results suggest that priority-based computation, long used to explain visual attention, may provide a structured and transparent framework for controlling the movements of artificial agents.

\section{Methods}\label{sec:methods}

\subsection{Visual search task and dataset}\label{sec:m-dataset}

Human eye-movement data came from a public dataset containing 114,232 first fixations collected from 333 participants across 11 experiments \citep{drennan_what_2026}. Participants searched for a predefined target (e.g., a green circle) among heterogeneous shapes, with a color-singleton distractor present on some trials. Target and distractor locations were randomized on each trial. Because the original stimulus images were unavailable, we reconstructed canonical displays from trial parameters, including set size, item colors, and target and distractor locations. We excluded trials containing motion, onset, or low-salience distractors, as well as one experiment in which the singleton color alternated across blocks. % TODO: cite the Gaspelin & Luck (2018) empirical paper (Exp. 4) from the Drennan & Gaspelin (2026) reference list.
We analyzed the first fixation on each trial as a direct measure of attentional priority. Participants were randomly assigned to training and held-out test sets using an 80/20 split.

\subsection{The priority-map model}\label{sec:m-search}

For each candidate item $i$, attentional priority was defined as a weighted sum of salience, goal, and selection history:
\begin{equation}
F(i) = w_{S}S(i) + w_{G}G(i) + w_{H}H(i)
\end{equation}
where $S(i)$ measures perceptual salience, $G(i)$ measures similarity to the target-defining features, and $H(i)$ reflects the influence of previous target locations. $w_{S}$, $w_{G}$, and $w_{H}$ are weights to modulate the influence of the respective maps.

\paragraph{Salience map} Raw salience is implemented as a simple color-contrast detector \citep{itti_model_1998,itti_computational_2001}. Specifically, $S(i)$ is the distance of item $i$'s color from the average color of all items, scaled to $[0, 1]$. A color singleton is salient because its color deviates from the majority.

\paragraph{Goal map} The goal map measures how well each item matches the search template \citep{duncan_visual_1989,wolfe_guided_2021}, and equally weights color (e.g., green) and shape (e.g., circle). A green circle matches on both and is strongly positive, whereas a red cross mismatches on both and is strongly negative. The map is constrained to be $[-1, 1]$, allowing for both upweighting of target features and downweighting of distractor features.

\paragraph{History map} Selection history was implemented as a leaky accumulator for each item location. After each trial, the memory trace at each item location decayed, and the memory trace at the current target location was updated:
\begin{equation}\label{eq:search-history}
H \leftarrow (1 - \eta_{H})H + \eta_{H}e
\end{equation}
where $e$ marks the current target location, $\eta_{H}$ controls the rate of updating, and $1 - \eta_{H}$ controls the rate of decay. A higher $\eta_{H}$, therefore, means the memory trace is more biased toward recent events.

\paragraph{Readout} The probability that the first fixation landed on item $i$ was given by a softmax over all item priorities:
\begin{equation}\label{eq:softmax}
P(\text{first fixation} = i) = \frac{\exp[F(i)]}{\sum_{j} \exp[F(j)]}
\end{equation}
Thus, higher-priority items had a higher landing probability. To generate a predicted fixation for a trial, we drew a single item from this distribution.

\paragraph{Fitting and evaluation} The model had four free parameters: the salience, goal, and history weights ($w_{S}$, $w_{G}$, and $w_{H}$) and the memory-update rate ($\eta_{H}$). We constrained $w_{G} > 0$ and $0 < \eta_{H} < 1$, while leaving $w_{S}$ and $w_{H}$ unconstrained. Parameters were estimated by maximum likelihood using the Adam optimizer for 300 full-batch iterations with a learning rate of 0.05.

We evaluated the fitted model on held-out participants using the negative log-likelihood (NLL) of the observed first-fixation locations. We also constructed a human-consistency baseline: for each display configuration (a unique combination of set size, target and singleton locations, and colors), we estimated fixation probabilities based on the training set and used them to compute NLLs for the test set. Model and baseline NLLs were averaged within participant and compared using a paired $t$-test, with Cohen's $d$ as the effect size.

\subsection{Behavioral-signature analyses}\label{sec:m-signatures}

To test whether the fitted model can reproduce key patterns observed in the literature, we generated model predictions for each held-out trial using actual trial parameters and trial history. We measured three behavioral signatures: (1) oculomotor suppression, defined as the difference in landing rates between the singleton and an average non-singleton distractor; (2) salience dependence, assessed in a counterfactual pass where predictions for distractor-present trials were generated assuming a low-salience singleton, and (3) target-location priming, defined as the difference in target-landing first fixations between trials on which the target location repeated versus changed. Participant-level landing probabilities were compared using paired $t$-tests with $\alpha = 0.05$ and Cohen's $d$ as the effect size.

\subsection{The priority-field model}\label{sec:m-field}

The priority-field model follows the same basic form as the search model. For each of 360 movement directions $q$ around the agent, movement priority is the sum of salience, goal, and history components:
\begin{equation}
F(q) = f_{S}(S(q)) + f_{G}(G(q)) + f_{H}(H(q))
\end{equation}

\paragraph{Distance psychophysics} Whereas the search model uses a hand-specified psychophysical transformation from stimulus to perceptual salience, an agent in a hazardous environment must learn its own psychophysical function to transform raw distance to perceived closeness. For each direction, distance $d(q)$ is passed through a sigmoid function, with its shape modulated by a learned sensitivity parameter, $k_{q}$:
\begin{equation}\label{eq:closeness}
c(q) = 2\bigl(1 - \sigma(k_{q}d(q))\bigr), \quad \sigma(z) = \frac{1}{1 + e^{-z}}
\end{equation}
This specification allows raw distance to be transformed into perceived closeness via a learned psychophysical function. It further allows the agent to learn to adjust the steepness of the psychophysical function for each movement direction.

\paragraph{Salience field} At each time $t$, the agent conducts a 360-degree distance scan and stores the distance to the nearest obstacle or arena wall for each direction at $t$ and $t-1$. Applying the psychophysics transformation yields previous and current closeness values, $c_{t-1}(q)$ and $c_{t}(q)$. These values define two salience channels for each direction:
\begin{equation}
\text{proximity: } c_{t}(q), \qquad \text{looming: } \bigl(c_{t}(q) - c_{t-1}(q)\bigr)\,c_{t}(q)
\end{equation}
A multilayer perceptron with one hidden layer of eight rectified linear units (ReLUs; $2\to8\to1$) combines the two channels into a coherent salience field. In other words,
\begin{equation}\label{eq:salience-field}
f_{S}(S(q)) = M_{S}(S(q)), \quad S(q) = \bigl[c_{t}(q),\ \bigl(c_{t}(q) - c_{t-1}(q)\bigr)\,c_{t}(q)\bigr]
\end{equation}

\paragraph{Goal field} The goal field encodes the goal's influence on each movement direction. The goal value for direction $q$ is the cosine of the angle between that candidate direction and the direction from the agent to the goal:
\begin{equation}\label{eq:goal}
G(q) = \cos(q - \theta_{G})
\end{equation}
where $\theta_{G}$ is the direction from the agent to the goal. The angular alignment term, $G(q)$, equals 1 when the candidate direction points straight at the goal, 0 when it is perpendicular to the goal, and $-1$ when it is opposite to the goal. Essentially, this is the movement version of target-feature matching in visual search. Furthermore, $G(q)$ is modulated by a mapping function:
\begin{equation}\label{eq:goal-field}
f_{G}(G(q)) = A_{G}\!\left(\frac{d_{G}}{R}\right) G(q)
\end{equation}
where $R$ is the sensing range, and $A_{G}$ is a multilayer perceptron with one hidden layer of eight rectified linear units ($1\to8\to1$). Compared to the search model, the scalar weight $w_{G}$ becomes a learned function of goal distance, $A_{G}(d_{G}/R)$.

\paragraph{History field} The history field used the same parameterization as in the search model:
\begin{equation}
f_{H}(H(q)) = w_{H}H(q)
\end{equation}
Goal history was represented by a leaky accumulator on an $8\times8$ spatial grid updated after each goal onset:
\begin{equation}\label{eq:grid-history}
W \leftarrow (1 - \eta_{H})W, \quad W_{\text{cell(goal)}} \leftarrow W_{\text{cell(goal)}} + \eta_{H}
\end{equation}
where $W$ is the $8\times8$ grid of accumulated history weights, $\eta_{H}$ is the update rate (and $1 - \eta_{H}$ the decay rate), and $\text{cell(goal)}$ is the grid cell containing the current goal. After each goal onset, every cell decays by a factor of $(1 - \eta_{H})$ and the cell in which the goal appeared is incremented by $\eta_{H}$.

To map the leaky accumulator to effects on movement directions, the grid was transformed to egocentric coordinates by treating each grid cell as a weak goal. Here we reused the goal field's gain function, $A_{G}$, and cosine geometry:
\begin{equation}\label{eq:history-field}
H(q) = \sum_{c} W_{c}\, A_{G}\!\left(\frac{d_{c}}{R}\right) \cos(q - \theta_{c})
\end{equation}
where $W_{c}$ is the accumulated history at grid cell $c$, and $d_{c}$ and $\theta_{c}$ are the grid cell's distance and angular alignment relative to the agent. The history field was active only when the current goal was not visible. Essentially, the history field exerts a pull force toward the memory-weighted previous goal locations when the agent lacks a visible goal.

\paragraph{Readout} The three fields are summed into a 360-dimensional priority field
\begin{equation}
\mathbf{F} = [F(q)]_{q = 1,\ldots,360}, \quad F(q) = f_{S}(S(q)) + f_{G}(G(q)) + f_{H}(H(q))
\end{equation}
A planner network maps this field to the movement,
\begin{equation}
\mathbf{a} = \mathrm{Planner}(\mathbf{F})
\end{equation}
where the planner is a multilayer perceptron with two hidden layers of 360 rectified linear units ($360\to360\to360\to2$). The movement vector $\mathbf{a} \in \mathbb{R}^{2}$ indicates the horizontal and vertical steps the agent should take. Because training does not fix the overall sign of the field, we report the field in the convention in which higher values favor movement in that direction.

\subsection{Reach--avoid task}\label{sec:m-task}

The agent operated continuously in a bounded $800\times800$-pixel arena containing 10 moving obstacles and a single goal destination. Each obstacle was a disk of the same size as the agent (20-pixel radius) with a randomly chosen speed of 1 to 3 pixels per step along the vertical and horizontal axes. Obstacles traveled diagonally at constant velocity, bounced off the arena walls, and passed freely through one another. The agent had a maximum speed of 10 pixels per step. A goal was counted as reached once the agent came within 30 pixels of its location. Reaching a goal or failing to reach it within 6 seconds spawned a new goal elsewhere in the arena. At each frame, the agent received the distance to the nearest obstacle or arena wall in each direction, for the current and previous frames, along with a vector to the goal.

\subsection{Training}\label{sec:m-training}

Expert demonstrations were generated by a scripted expert that moved to the goal and avoided nearby obstacles. The expert played 20 independent episodes of the continuous task (10 obstacles, unbiased goal locations), yielding 12,000 state-action pairs in the demonstration dataset.

We trained the agent via behavioral cloning by minimizing the mean squared error between its predicted movements and the expert's movements. We used the Adam optimizer with a base learning rate of $10^{-3}$ and a batch size of 64 for 500 epochs. We employed a plateau-based learning-rate schedule to jointly optimize the distance sensitivity $k$, the salience network $M_{S}$, the goal-gain network $A_{G}$, and the movement planner. Because the expert was memoryless and its demonstrations therefore contained no history-driven behavior, we fixed $w_{H} = 0$ during training and later set $w_{H}$ and $\eta_{H}$ separately (see Statistical Learning).

\subsection{Alternative models}\label{sec:m-baselines}

We compared the priority-based model with two alternative models that received the same input, but without an explicit priority-based architecture. The multilayer perceptron (MLP) baseline had two hidden layers with 360 and 180 rectified linear units ($722\to360\to180\to2$). The transformer baseline embedded each input value as a 16-dimensional token, added positional encodings, and applied a two-layer encoder with four attention heads and a feedforward width of 64. Each token's representation was then averaged across the embedding dimension, and the resulting 722-length vector was mapped to the movement by a linear layer. Both baselines were trained on the same demonstrations using the same training regime.

\subsection{Evaluation}\label{sec:m-evaluation}

Trained agents were evaluated on different variants of the continuous reach--avoid task. First, we tested model performance on the trained case, where the testing scenarios used the same configuration as during training. Second, we tested the agents' performance in novel, more challenging scenes with 20, 30, 40, and 50 obstacles. Obstacle speed was held at the training level. Third, we tested performance in scenarios with obstacle speeds at 1.5$\times$, 2$\times$, 2.5$\times$, 3$\times$, 3.5$\times$, and 4$\times$ of the training level, while keeping the number of obstacles at the training level. Each scenario ran continuously for 6,000 simulation steps (two minutes at 50 steps per second).

Evaluation metrics include goals reached per minute and collisions per minute, which quantify speed and safety, respectively. Each model was evaluated on each scenario across 20 random seeds. We report the mean values across seeds with bootstrapped 95\% confidence intervals. Statistical tests were performed using paired $t$-tests with $\alpha = 0.05$ and Cohen's $d$ to measure effect size.

\subsection{Statistical learning}\label{sec:m-statlearn}

We tested whether the history mechanism enabled the agent to learn and respond to spatial regularities. Each trial began with the agent at the center of an arena containing 10 obstacles but no visible goal. This goal-free period allowed learned anticipatory movement to emerge. After 50 steps (1 second), a goal appeared in one of two symmetric 90-degree wedges to the left or right, at 250--330 pixels from the arena center. During biased trials, the goal appeared on the left with 70\% probability; during unbiased trials, it appeared on either side with equal probability. The agent returned to the center either upon reaching the goal or after a 6-second timeout.

We equipped the trained agent with grid memory ($w_{H} = 0.2$, $\eta_{H} = 0.15$) and tested it on 90 biased trials followed by 90 unbiased trials. We repeated the procedure for 20 runs with different random seeds. As a baseline, we tested the same agent with memory disabled ($w_{H} = 0$) using the same seeds. Anticipatory movement was measured as the agent's horizontal offset from the arena center when the goal appeared. Negative values thus indicate movement toward the high-probability region. For each seed, we averaged the offset for biased and unbiased trials, and compared the memory and baseline agents using a paired $t$-test with $\alpha = 0.05$ and Cohen's $d$ as the effect size. For visualization, we also averaged the offset within each consecutive 10-trial bin and computed 95\% bootstrap confidence bands.

Finally, we tested whether the history parameters could be recovered from behavior. Holding all network weights fixed, we estimated only $w_{H}$ and $\eta_{H}$ by minimizing mean squared error between predicted and recorded goal-free movements using Adam (learning rate 0.05, 400 iterations). The fitted parameters recovered the generating values to two decimal places.

\bibliographystyle{elsarticle-harv}
\bibliography{references}

\end{document}